\documentclass[10pt,twocolumn,letterpaper]{article}

\usepackage{wacv}
\usepackage{times}
\usepackage{epsfig}
\usepackage{graphicx}
\usepackage{amsmath}
\usepackage{amssymb}

\wacvfinalcopy 

\ifwacvfinal
\usepackage[breaklinks=true,bookmarks=false]{hyperref}
\else
\usepackage[pagebackref=true,breaklinks=true,colorlinks,bookmarks=false]{hyperref}
\fi

\begin{document}

\title{Coarse-to-Fine Gaze Redirection with Numerical and Pictorial Guidance}

\author{Jingjing Chen$^1$,\, Jichao Zhang$^2$,\, Enver Sangineto$^2$,\,
Tao Chen$^3$\footnotemark[1],\,
Jiayuan Fan$^4$,\, Nicu Sebe$^{2,5}$\\
$^1$Zhejiang University, $^2$University of Trento\\
$^3$School of Information Science and Technology, Fudan University\\
$^4$Academy for Engineering and Technology, Fudan University, $^5$Huawei Research Ireland}



\maketitle
\renewcommand{\thefootnote}{\fnsymbol{footnote}}
\thispagestyle{empty}
\footnotetext[1]{Tao Chen is the corresponding author.}

\begin{abstract}
Gaze redirection aims at manipulating the  gaze of a given face image  with respect to a desired direction (i.e., a reference angle) and it can be applied to many real life scenarios, such as video-conferencing or taking group photos. However, previous work on this topic mainly suffers of two limitations: (1) Low-quality image generation and (2) Low redirection precision. In this paper, we propose to alleviate these problems by means of a novel gaze redirection framework which exploits both a numerical and a pictorial direction guidance, jointly with a coarse-to-fine learning strategy. Specifically, the coarse branch learns the spatial transformation which warps input image according to desired gaze. On the other hand, the fine-grained branch consists of a generator network with conditional residual image learning and a multi-task discriminator.  This second branch reduces the gap between the previously warped image and the ground-truth image and recovers finer texture details. Moreover, we propose a numerical and pictorial guidance module~(NPG) which uses a pictorial gazemap description and numerical angles as an extra guide to further improve the precision of gaze redirection. Extensive experiments on a benchmark dataset show that the proposed method outperforms the state-of-the-art approaches in terms of both image quality and redirection precision. The code is available at \url{https://github.com/jingjingchen777/CFGR}
\end{abstract}

\section{Introduction}

Gaze redirection is a new research topic in computer vision and computer graphics and its goal is to manipulate the eye region of an input image, by changing the gaze according to a reference angle. This task is important in many real-world scenarios. For example, when taking a group photo, it rarely happens that everyone is simultaneously looking at the camera, and adjusting each person's gaze with respect to the same direction (e.g., the camera direction) can make the photo look better and user acceptable. In another application scenario, when talking in a video conferencing system, eye contact is important as it can express attentiveness and confidence. However, due to the location disparity between the video screen and the camera, the participants do not have direct eye contact. Additionally, gaze redirection tasks can be applied to improve few-shot gaze estimation~\cite{yu2019improving,yu2019unsupervised} and domain transfer~\cite{Kaur_2020_WACV}.


Traditional methods are based on a 3D model which re-renders entire input region~\cite{banf2009example,wood2018gazedirector}. These methods suffer from two major problems: (1) it is not easy to render the entire input region and (2) they require an heavy instrumentation. Another type of gaze redirection is based on machine learning for image re-synthesis, such as DeepWarp~\cite{ganin2016deepwarp} or PRGAN~\cite{he2019photo}. DeepWarp~\cite{ganin2016deepwarp} employs a neural network to predict the dense flow field which is used to warp the input image with respect to the  gaze redirection. However, this method cannot generate perceptually plausible samples, as only using the pixel-wise differences between the synthesized and ground truth images is insufficient. PRGAN~\cite{he2019photo} proposes a GAN-based autoencoder with a cycle consistent loss for monocular gaze redirection and it can synthesize samples with high quality and redirection precision. However, its single-stage learning causes the corresponding appearance to look asymmetric. Overall, the previous results are still far from the requirements imposed by many application scenarios.


\begin{figure*}[ht]
\begin{center}
\includegraphics[width=1.0\linewidth]{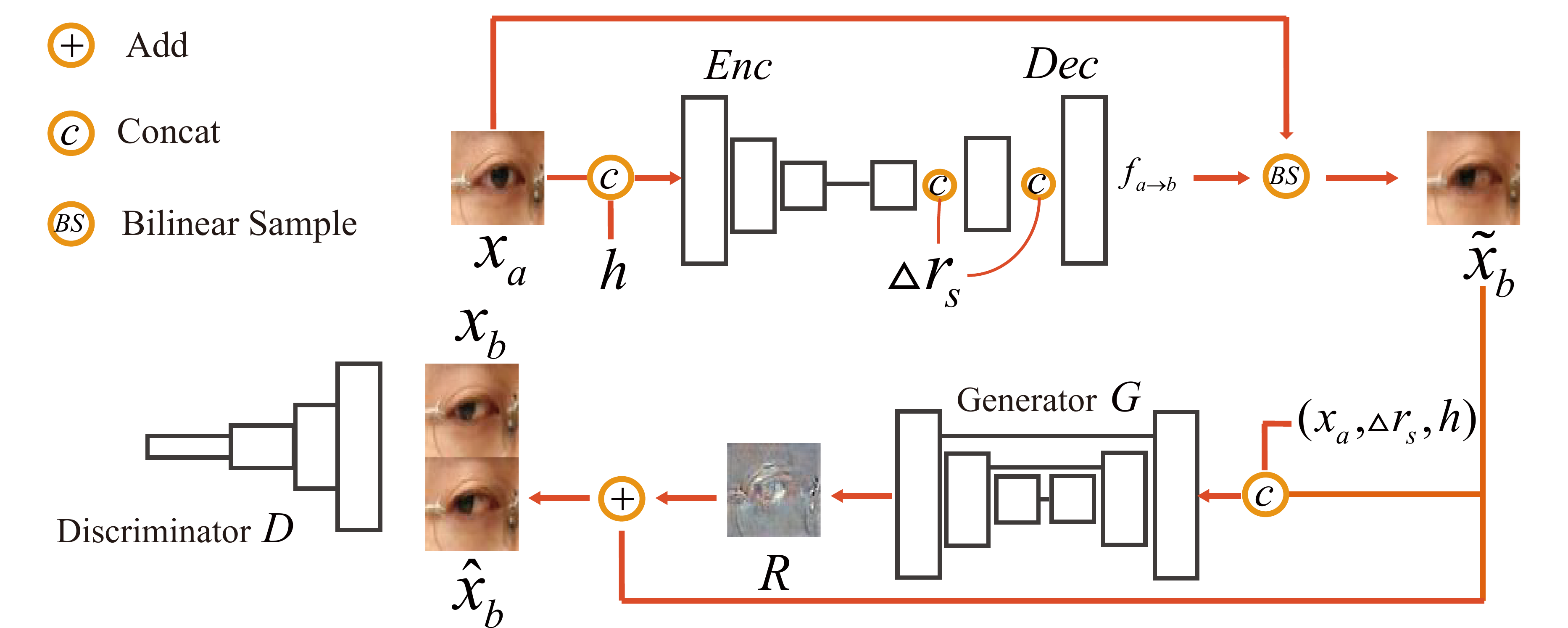}
\end{center}
\vspace{-0.5cm}
\caption{The pipeline of the proposed gaze redirection approach. The upper branch outputs a coarse-grained result $\tilde x_{b}$. The encoder $Enc$ takes as input the eye region $x_{a}$ and the head pose $h$, while the decoder $Dec$ takes as input the encoder latent code and $\triangle r_{s}$~(provided by the NPG module, not shown in the figure). The lower branch outputs fine-grained, final results. The generator $G$ outputs the residual image $R$, which is added to $\tilde x_{b}$. The refined results $\hat x_{b}$ and the ground truth $x_{b}$ are fed to the discriminator $D$.}
\label{fig:model}
\vspace{-0.3cm}
\end{figure*}

In this paper,
 we propose a coarse-to-fine strategy and we combine flow learning with adversarial learning to produce higher quality and more precise redirection results. As shown in Fig.~\ref{fig:model}, our model consists of three main parts. The first one is a coarse-grained model which is an encoder-decoder architecture with flow learning and models the eye spatial transformation. Specifically, this network is fed with source images and with the angle-difference vector between target and source. Second, in order to refine the warped results, we propose to use a conditional architecture, in which  the generator learns the residual image between the warped output and the ground truth. The goal of the generator is to reduce possible artifacts in the warped texture and the distortions in the eye shape. Finally, a discriminator network with gaze regression learning is used to ensure that the refined results have the same distribution and the same gaze angles as the ground truth. Additionally, we propose an NPG module which integrates the pictorial gazemap representation with numerical angles to guide the synthesis process. The intuitive idea is that the gazemap pictorial representation can provide additional spatial and semantic information of the target angle~(shown in
Fig. \ref{fig:model2}).

The main contributions of our work are:
\begin{enumerate}
\item We propose a coarse-to-fine eye gaze redirection model which combines flow learning and adversarial learning.
\item We propose an NPG module which integrates the pictorial gazemap with numerical angles to guide the generation process.
\item We present a comprehensive experimental evaluation demonstrating the superiority of our approach in terms of both image quality of the eye region and angle redirection precision.
\end{enumerate}

\section{Related Work}

{\bfseries Facial Attribute Manipulation}, an interesting multi-domain image-to-image translation problem, aims at modifying the semantic content of a facial image according to a specified attribute, while keeping other irrelevant regions unchanged.
Most works~\cite{choi2018stargan,zhang2018sparsely,liu2019stgan,perarnau2016invertible,Pumarola_ijcv2019, wu2020cascade, cheng2019tc, zhang2017st, he2019attgan, zheng2020survey, 8756586,Tang2020MultiChannelAS,Zheng2020ASO} are based on GANs and have achieved impressive facial attribute manipulation results. However, these methods tend to learn the style or the texture translation and are not good in obtaining high-quality, natural geometry translations. To alleviate this problem, Yin, et al.~\cite{yin2019geogan} proposed a geometry-aware flow which is learned using a geometry guidance obtained by facial landmarks. Wu, et al.~\cite{wu2019attribute} also exploits the flow field to perform spontaneous motion, achieving higher quality facial attribute manipulation. Eye gaze redirection can be considered as a specific type of facial attribute manipulation. To the best of our knowledge, our model is the first combining flow learning and adversarial learning for gaze redirection.

{\bfseries Gaze Redirection.} Traditional methods are based on a 3D model which re-renders the entire input region. Banf and Blanz~\cite{banf2009example} use an example-based approach to deform the eyelids and slides the iris across the model surface with texture-coordinate interpolation. GazeDirector~\cite{wood2018gazedirector} models the eye region in 3D to recover the shape, the pose and the appearance of the eye. Then, it feeds an acquired dense flow field corresponding to the eyelid motion to the input image to warp the eyelids. Finally, the redirected eyeball is rendered into the output image.

Recently, machine learning based methods have shown remarkable results using a large training set labelled with eye angles and head pose information. Kononenko and Lempitsky~\cite{kononenko2015learning} use random forests as the supervised learners to predict the eye flow vector for gaze correction. Ganin et al.~\cite{ganin2016deepwarp} use a deep convolution network with a coarse-to-fine warping operation to generate redirection results. However, these warping methods based on pixel-wise differences between the synthesized and ground-truth images, have difficulties in generating photo-realistic images and they fail in the presence of large redirection angles, due to dis-occlusion problems. Recently, PRGAN~\cite{he2019photo} adopted a GAN-based model with a cycle-consistent loss for the gaze redirection task and succeeded in generating better quality results. However these results are still far from being satisfactory. To remedy this, Zhang, et al.~\cite{zhang2020dual} developed a dual inpainting module, to achieve high-quality gaze redirection in the wild by interpolating the angle representation. However, also this method fails to redirect gaze with arbitrary angles.

Compared to the previous methods, our approach exploits a coarse-to-fine learning process and it learns the flow field  for the spatial transformation. This is combined  with adversarial learning to recover the finer texture details. Moreover, we are the first to propose utilizing the gaze map (i.e., the pictorial gaze representation) as an input to provide extra spatial and semantic information for gaze redirection. Empirically, we found that this is beneficial in order to improve the redirection precision.

\section{Method}
\begin{figure}[t]
\begin{center}
\includegraphics[width=1.0\linewidth]{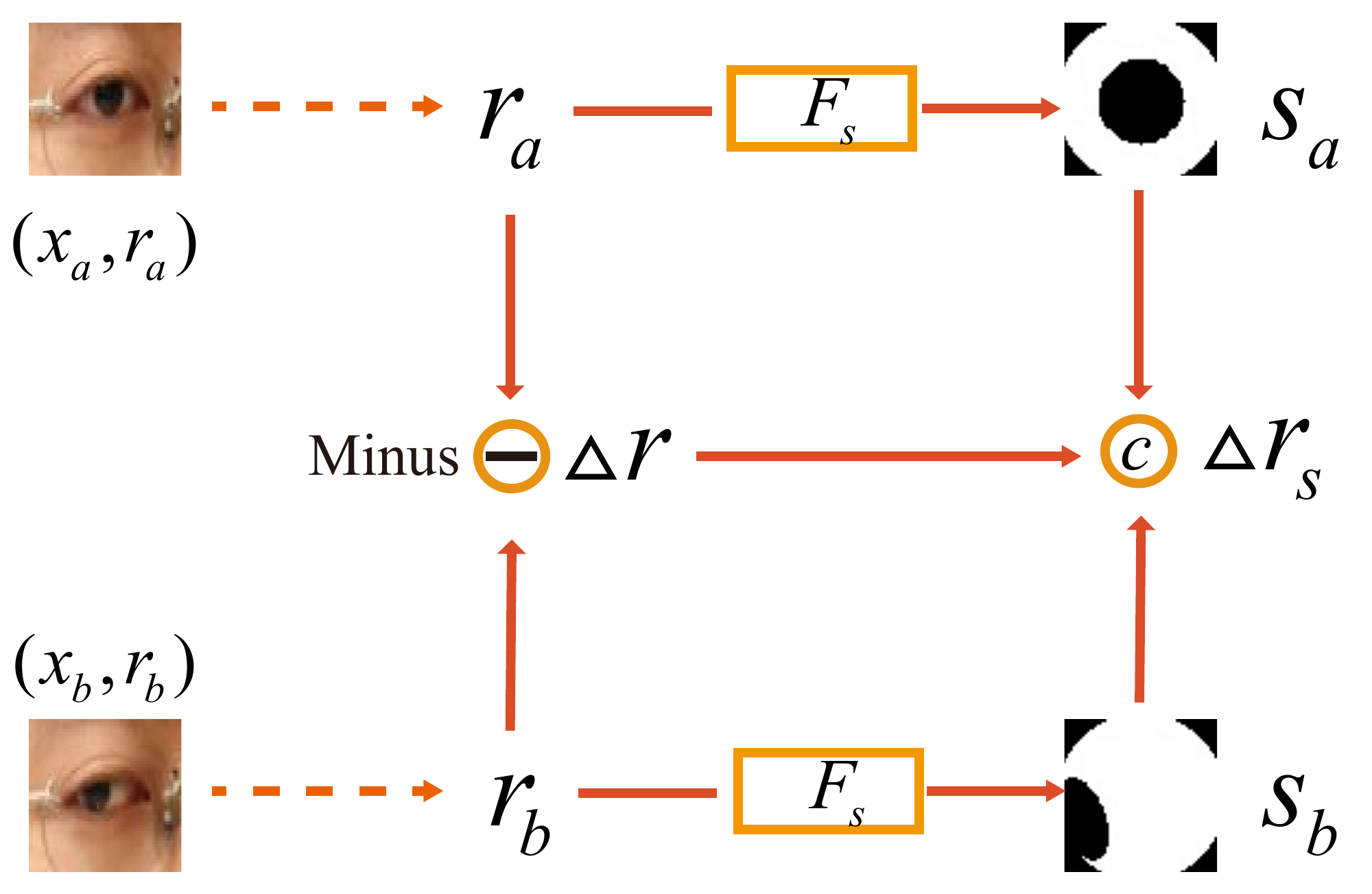}
\end{center}
\vspace{-0.5cm}
\caption{A scheme of the Numerical and Pictorial Guidance Module~(NPG). The angle difference vector $\triangle r$ is concatenated with two gazemaps [$S_{a}$, $S_{b}$] to get $\triangle r_{s}$. $S_{a}$ and $S_{b}$ correspond to the input angles $r_{a}$ and the target angles $r_{b}$, respectively. Note that the gazemap has a dimension different from the numeric angle $r$, thus a scale normalization is necessary. $F_{s}$ is the graphic tool producing the pictorial representation.}
\label{fig:model2}
\vspace{-0.2cm}
\end{figure}

\begin{sloppypar}
The pipeline of the proposed method is shown in Fig.~\ref{fig:model}. It is mainly split into two learning stages. In the coarse learning stage, an encoder-decoder architecture is proposed to generate coarse-grained results by learning the flow field necessary to warp the input image. On the other hand, the fine learning stage is based on a multi-task conditional GAN, in which  a generator  with conditional residual-image learning  refines the coarse output and recovers  finer texture details, eliminating the distortion in the eye geometry. Moreover, we propose an NPG module to guide both the coarse and the fine process (see Fig~\ref{fig:model2}). Before introducing the details, we first clarify the adopted notations.
\end{sloppypar}

$\bullet$ Two angle domains: source domain $A$ and target domain $B$. Note that paired samples exist in the two domains.

$\bullet$ $(x_{a}, r_{a}) \in A$ indicates the input eye image $x_{a} \in R^{m \times n \times c}$ from domain A and its corresponding angle pair $r_{a} \in R^{2}$, representing the eyeball pitch and yaw $[\theta, \phi]$. $(x_{b}, r_{b}) \in B$ are defined similarly. With $m,n,c$ we indicate, respectively, the width, the height and the channel number of the eye image. $x_{a}$ and $x_{b}$ are paired samples with different labeled angles. Our model learns the gaze redirection from $A$ to $B$.


$\bullet$ $\triangle r$ denotes the  angle vector difference between an input sample in  $A$ and the corresponding target sample in  $B$.

$\bullet$ $S \in R^{m \times n \times 2}$ denotes the two channel gazemap  pictorial representation (the eyeball and the iris), which is generated from an angle pair $r$. $S=F_{s}(r)$, where $F_{s}$ is a straightforward graphics tool used to generate pictorial representations of the eyeball and the iris, as shown in Fig.~\ref{fig:model2}. Note that each instance $x_{a}$ with same angle from domain $A$ has the same $S$.

\subsection{Flow-Field Learning for Coarse-Grained Gaze Redirection}

To redirect $x_{a}$ with an angle pair $r_{a}$ from domain $A$ to domain $B$, our encoder $Enc$ takes both $x_{a}$ and the corresponding head pose $h$ as inputs. Then,  the decoder  $Dec$  generates a coarse-grained output using both the encoded code and $\triangle r_{s}$ (provided by the NPG, see later). As shown in Fig.~\ref{fig:model}, $\triangle r_{s}$ is concatenated into different scales of $Dec$ to strengthen the guided ability of the conditional information. This can be formulated as follows:
\begin{equation}
\begin{aligned}
f_{a \rightarrow b} = Dec(Enc(x_{a}, h), \triangle r_{s}),
\end{aligned}
\end{equation}

\noindent where $f_{a \rightarrow b}$ is the learned flow field from $x_{a}$ to $x_{b}$. Similarly to DeepWarp~\cite{ganin2016deepwarp}, we generate the flow field to warp the input image.
In more details, the last convolutional layer of $Dec$ produces a dense flow field~(a two-channel map) which is used to warp the input image $x_{a}$ by means of a bilinear sampler $BS$. Here, the sampling procedure samples the pixels of $x_{a}$ at pixel coordinates determined by the flow field $f_{a \rightarrow b}$:

\begin{equation}
\begin{aligned}
\tilde x_{b}(i, j, c) = x_{a}\{i + f_{a \rightarrow b}(i, j, 1), j + f_{a \rightarrow b}(i, j, 2),c\},
\end{aligned}
\end{equation}
\noindent where $\tilde x_{b}$ is the warped result representing the coarse output, $c$ denotes the channel of the image, and the curly brackets represent the bilinear interpolation which skips those positions with illegal values in the warping process. We use the $L2$ distance between the output $\tilde x_{b}$ and the ground truth $x_{b}$ as the objective function which is defined as follows:

\begin{equation}
\begin{aligned}
L_{recon} = \mathbb{E}[\Vert \tilde x_{b} - x_{b} \Vert _{2}]
\end{aligned}
\end{equation}

{\bfseries NPG with Gazemap.} As shown in Fig.~\ref{fig:model2}, we use the NPG output as an additional condition of the generation process. Jointly with the numerical gaze angle representation $r_{a}$, the pictorial gazemap $S$  is concatenated in a multimodal term to provide additional spatial and semantic information about the angle direction.

First, and differently from previous work in gaze redirection~\cite{ganin2016deepwarp,he2019photo}, we compute the angle vector difference $\triangle r=r_{b} - r_{a}$, which is used as input instead of the absolute target angle to better preserve identity, similarly to~\cite{wu2019relgan}. Next, we generate the corresponding gazemap $S$ of the angles $r_{a}$ and $r_{b}$ by means of a synthesis process $F_{s}$ (details can be found below). Then, we concatenate $S_{a}$, $S_{b}$ and $\triangle r$ into a single term to get $\triangle r_{s}$:

\begin{equation}
\begin{aligned}
\triangle r_{s} = [\triangle r, S_{a}, S_{b}].
\end{aligned}
\end{equation}

We detail below how we generate the gazemap ($F_{s}$). As shown in~\cite{park2018deep}, our gazemap is also a two-channel Boolean image: one channel is for the eyeball which is assumed to be a perfect sphere, and the other channel is for the iris which is assumed to be a perfect circle. For an output map of size $m \times n$, with the projected eyeball diameter $2k=1.2n$, the coordinates $(\mu, \nu)$ of the iris center can be computed as follows:

\begin{equation}
\begin{aligned}
&\mu = \frac{m}{2} - k \cos\left(\arcsin \frac{1}{2}\right) \sin\phi\cos\theta  \\
&\nu = \frac{n}{2} - k \cos\left(\arcsin \frac{1}{2}\right)\sin\theta,
\end{aligned}
\end{equation}
where the input gaze angle is $r=(\theta, \phi)$. The iris is drawn as an ellipse with the major-axis diameter of $k$, and the minor-axis diameter of $r\left|\cos\theta\cos\phi\right|$. Note that the synthesized gazemap only represents the gaze angle, without identity details of the specific eye  sample. 

\subsection{Multi-task cGAN for Fine-grained Gaze Redirection}
The warped result is inevitably blurry when using only the $L_{2}$ loss. Additionally, it also suffers from unwanted artifacts and unnatural distortions in the shape of the iris for large redirection angles. To remove these problems, we employ a generator $G$ to refine the output of the decoder. Instead of manipulating the whole image directly, we use $G$ to learn the corresponding residual image $R$, defined as the difference between the coarse output and the ground-truth. In this way, the manipulation can be operated with modest pixel modifications which provide high-frequency details, while preserving the identity information of the eye shape. The learned residual image is added to the coarse output of  $Dec$:
\begin{equation} \label{residual_learning}
\begin{aligned}
\hat x_{b} = R + \tilde x_{b}.
\end{aligned}
\end{equation}
where $\hat x_{b}$ represents the refined output.

{\bfseries Conditional Residual Learning.} Learning the corresponding residual image $R$ is not a simple task as it requires the generator to be able to recognize subtle differences. Additionally, previous works~\cite{zhu2017unpaired, ganin2016deepwarp} indicate that introducing a suitable conditional information improves the performance of $G$. For this reason, we employ the input image $x_{a}$ and the head pose $h$ as conditional inputs for $G$. We also take the NPG output $\triangle r_{s}$ as input to provide stronger conditional information. The conditional residual image learning phase can be written as:
\begin{equation}
\label{conditional information}
\begin{aligned}
R = G(\tilde x_{b}, x_{a},h,\triangle r_{s}).
\end{aligned}
\end{equation}
Similarly to the coarse process, the image reconstruction loss,
based on the $L2$ distance, is defined as follows:
\begin{eqnarray}
L_{g\_recon} = \mathbb{E}\left[\Vert \hat x_{b} - x_{b} \Vert _{2}\right].
\label{image reconstruction loss}
\end{eqnarray}

The $L2$ loss  penalizes pixel-wise discrepancies but it usually causes blurry results. To overcome this issue, we adopt the perceptual loss proposed in~\cite{johnson2016perceptual}. We use a VGG-16 network~\cite{simonyan2014very}, pre-trained on ImageNet~\cite{deng2009imagenet}, which we denote  as $\Phi$. The perceptual loss is defined as follows:

\begin{equation}
\label{per loss}
\begin{aligned}
 L_{g\_per} &=&  \mathbb{E}\left[\frac{1}{h_{j}w_{j}c_{j}}\Vert \Phi_{j}(\hat x_{b}) - \Phi_{j}(x_{b}) \Vert _2\right] \nonumber \\
  &+& \mathbb{E}\left[\sum^{J}_{j=1}\Vert \Psi_{j}(\hat x_{b}) - \Psi_{j}(x_{b}) \Vert _2\right],
\end{aligned}
\end{equation}
where $\Phi_{j}( \cdot) \in \mathbb{R}^{h_{j}\times w_{j} \times c_{j}}$ is the output of the $j$-th layer of $\phi$. In our experiments, we use the activation of the 5th layer. $\Psi_{j}$ denotes the Gram matrix (for more  details we refer the reader to~\cite{gatys2016image}).


{\bfseries Multi-task Discriminator Learning.} We use a multi-task discriminator in our model. Different from $G$, which is conditioned using multiple terms, the discriminator $D$ does not use them as input. Moreover, $D$ not only performs adversarial learning~($D_{adv}$)  but also regresses the gaze angle~($D_{gaze}$). Note that $D_{adv}$ and $D_{gaze}$ share most of the layers with the exception of the last two layers. The regression loss is defined as follows:
\begin{equation}
\label{gaze estimation loss}
\begin{aligned}
L_{d\_gaze} &= \mathbb{E}\left[\Vert D_{gaze}( x_{b}) - r_{b} \Vert _{2}\right] \\
L_{g\_gaze} &= \mathbb{E}\left[\Vert D_{gaze}( \hat x_{b}) - r_{b} \Vert _{2}\right].
\end{aligned}
\end{equation}
The adversarial loss for $D$ and $G$ is defined as:
\begin{equation}
\label{eq_gan_loss}
\begin{aligned}
\mathop{min}\limits_{G}\mathop{max}\limits_{D}L_{adv}&= \mathbb{E}\left[logD_{adv}(x_{b}) \right]\\
&+ \mathbb{E}\left[log(1-D_{adv}(\hat x_{b}))\right].
\end{aligned}
\end{equation}

\begin{figure*}[htp]
\begin{center}
\includegraphics[width=1.0\linewidth]{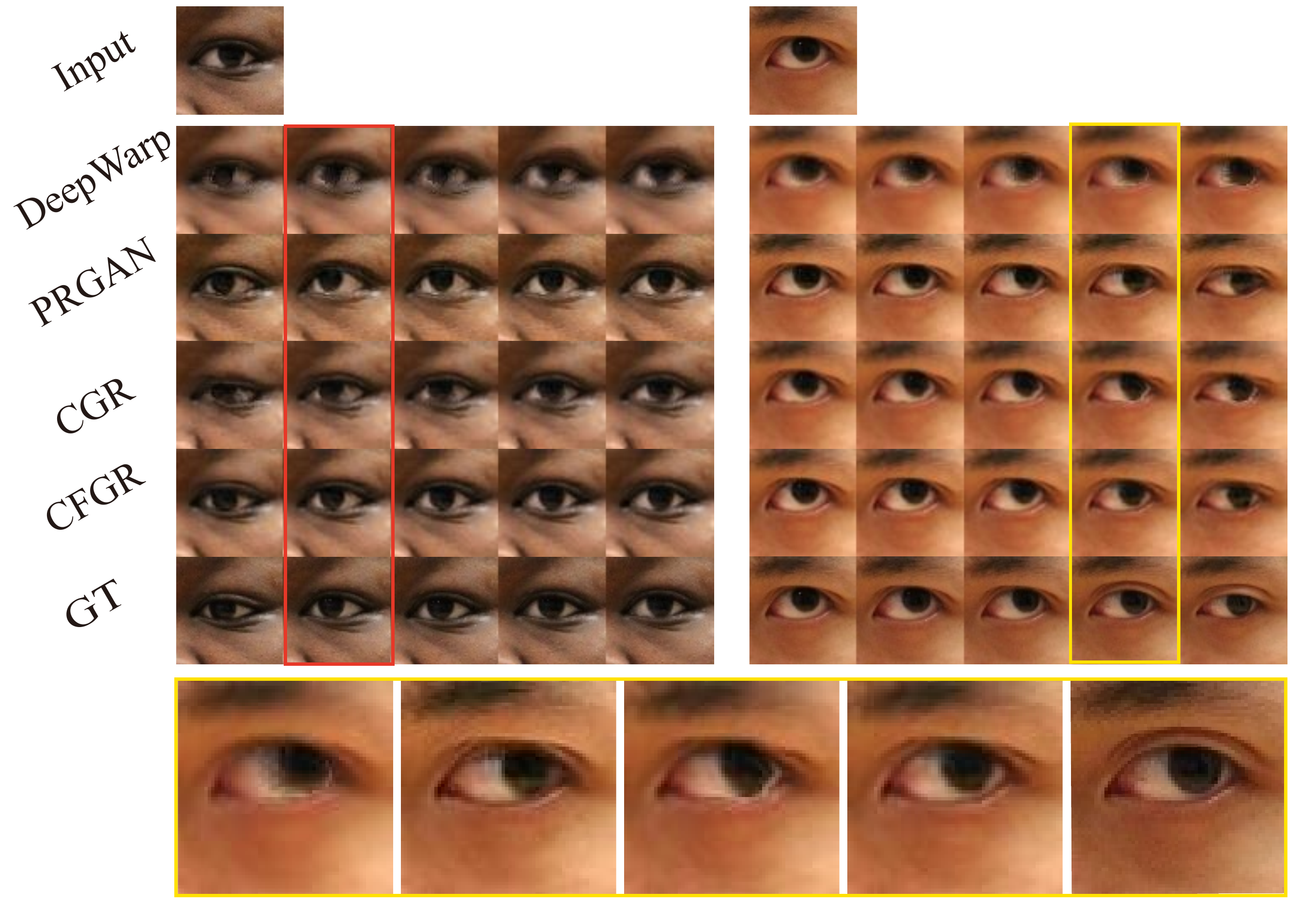}
\end{center}
\vspace{-0.5cm}
\caption{A qualitative comparison of different methods using redirection results with 10 different target angles~($\pm$ 15$^\circ$ head pose). The last row shows a magnification of the details marked with a yellow box in the previous rows, which  correspond, from left to right, to  DeepWarp, PRGAN, CGR, CFGR and GT.}
\label{fig:exp1_quan}
\vspace{-0.4cm}
\end{figure*}

\begin{sloppypar}
{\bfseries Overall Objective Functions.} As aforementioned, we use $L_{recon}$ to train the encoder-decoder $Enc$ and $Dec$ to get the coarse-grained results. The overall objective function for $D$ is:
\end{sloppypar}
\begin{equation}
\begin{aligned}
L_{D} = \lambda_{1}L_{d\_gaze} - L_{adv}.
\end{aligned}
\end{equation}
The overall objective function for $G$ is:
\begin{equation}
\begin{aligned}
L_{G} = \lambda_{2}L_{g\_recon} + \lambda_{3}L_{g\_per} +\lambda_{4}L_{g\_gaze}+L_{adv}.
\end{aligned}
\end{equation}
$\lambda_{1}$ $\lambda_{2}$, $\lambda_{3}$ and $\lambda_{4}$ are hyper-parameters controlling the contributions of each loss term. Note that $L_{G}$ is used only to optimize $G$, but not to update  $Enc$ and $Dec$.

\section{Experiments}

We first introduce the dataset used for our evaluation, the training details, the baseline models and the adopted metrics. We then compare the proposed model with two baselines using both a qualitative and a quantitative analysis. Next, we present an ablation study to demonstrate the effect of each component in our model, e.g., flow learning, residual image learning and the NPG module. Finally, we investigate the efficiency of our model. We refer to the full model as CFGR, and to the encoder-decoder with the only coarse-grained branch as CGR.

\begin{figure*}[t]
\begin{center}
\includegraphics[width=1.0\linewidth]{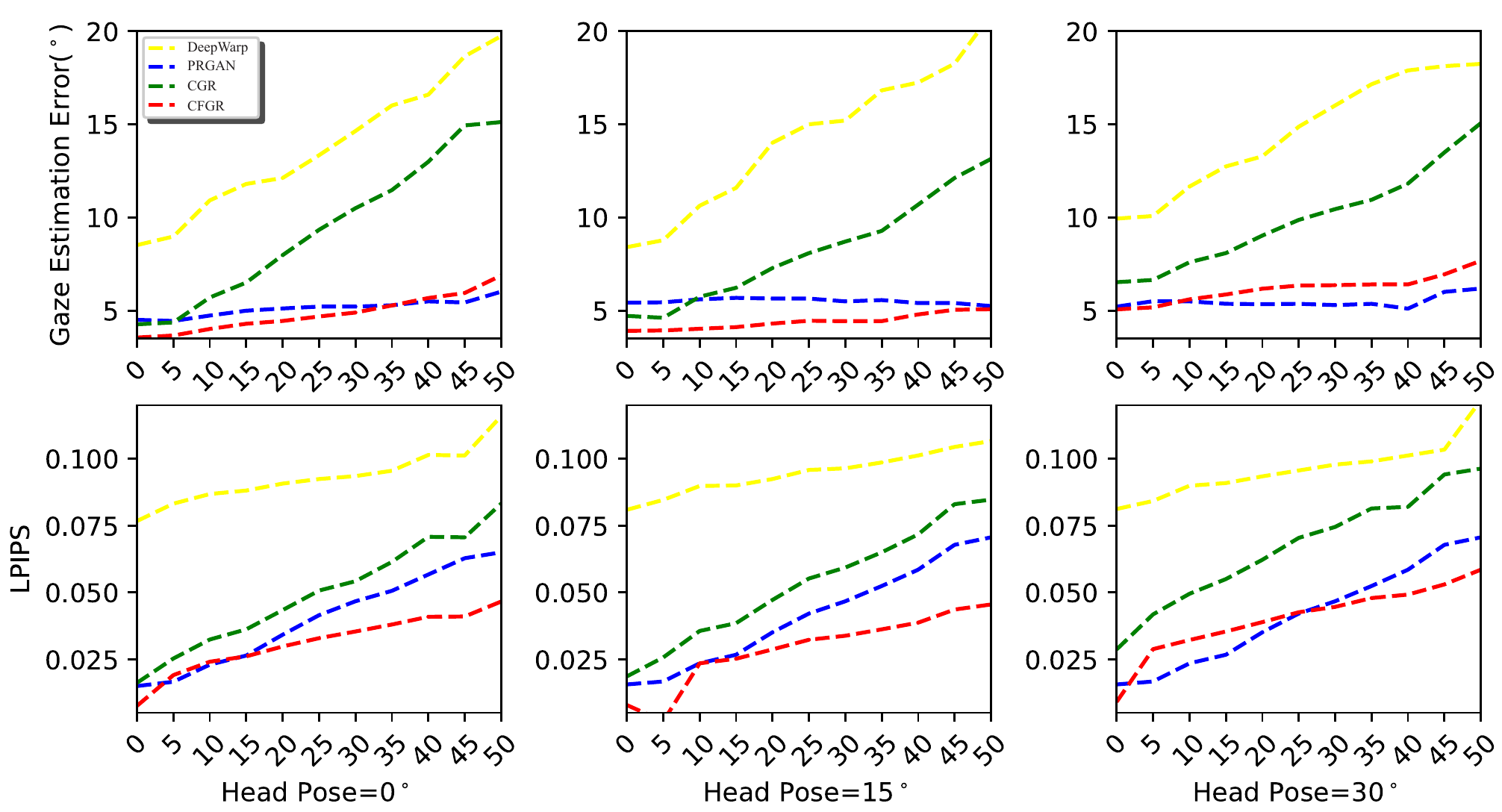}
\end{center}
\vspace{-0.5cm}
\caption{A quantitative evaluation of the gaze redirection results using three classes of head pose. First row: gaze estimation error. Second row: LPIPS scores. Lower is better for both  metrics. Note that we combine the results of $\pm 15^\circ$ and $\pm 30^\circ$ head poses into $15^\circ$ and $30^\circ$.}
\label{fig:exp1}
\vspace{-0.3cm}
\end{figure*}

\subsection{Experimental Settings}
{\bfseries Dataset.} We use the Columbia gaze dataset~\cite{Smith:2013:GLP:2501988.2501994}, containing 5,880 images of 56 persons with varying gaze directions and head poses. For each subject, there are 5 head directions (\hyphenation{hyphen-ation}{$[-30^\circ,-15^\circ,0^\circ \-,15^\circ,30^\circ]$}) and 21 gaze directions ($[-15^\circ,-10^\circ,-5^\circ,$ $0^\circ,5^\circ,10^\circ,15^\circ]$ for the yaw angle and $[-10^\circ, 0^\circ, 10^\circ]$ for the pitch angle, respectively). In our experiments, we use the same dataset settings of PRGAN~\cite{he2019photo}. In details, we use a subset of 50 persons~(1-50) for training and the rest~(51-56) for testing. To extract the eye region from the face image, we employ an external face alignment library (dlib~\cite{king2009dlib}). Fixed $64 \times 64$ image patches are cropped as the input images for both training and testing. Both the RGB pixel values  and the gaze directions are normalized in the range $[-1.0, 1.0]$. Other publicly available gaze datasets, e.g., MPIIGaze~\cite{zhang2017mpiigaze} or EYEDIAP~\cite{funes2014eyediap}, provide only low-resolution images and have not been considered in this evaluation.

{\bfseries Training Details.} CGR is trained independently of the generator and the discriminator and it is optimized firstly, followed by $D$ and $G$. We use the Adam optimizer  with $\beta_{1}=0.5$ and $\beta_{2}=0.999$. The batch size is 8 in all the experiments. The learning rate for CGR is 0.0001. The learning rate for $G$ and $D$ is 0.0002  in the first 20,000 iterations, and then it is linearly decayed to 0 in the remaining iterations. $\lambda_{1}=5$, $\lambda_{2}=0.1$, $\lambda_{3}=100$ and $\lambda_{4}=10$ in our all experiments.

{\bfseries Baseline Models.} We adopt DeepWarp~\cite{ganin2016deepwarp} and PRGAN~\cite{he2019photo} as the baseline models in our comparison. We use the official code of PRGAN\footnote{https://github.com/HzDmS/gaze\_redirection} and train it using the default parameters. We reimplemented DeepWarp, as its code is not available. In details, different from the original DeepWarp, which is used only for a gaze redirection task with a single direction, we trained DeepWarp for gaze redirection tasks in arbitrary directions. Moreover, DeepWarp uses 7 eye landmarks as input, including the pupil center. However, detecting the pupil center is very challenging. Thus, we computed the geometric center among the 6 points as a rough estimation of the pupil center.

{\bfseries Metrics.} How to effectively evaluate the appearance consistency and the redirection precision of the generated images is still an open problem. Traditional metrics, e.g., PSNR and MS-SSIM, are not correlated with the perceptual image quality~\cite{zhang2018unreasonable}. For this reason, and similarly to PRGAN, we adopted the LPIPS metric~\cite{zhang2018unreasonable} to compute the perceptual similarity in the feature space and evaluate the quality of redirection results. Moreover, we use GazeNet~\cite{zhang2017mpiigaze} as our gaze estimator and we pre-trained GazeNet on the MPIIGaze dataset to improve its gaze estimation. 

\begin{table}
\begin{center}
\caption{Gaze-redirection quantitative evaluation. The scores represent the average of three head poses over ten redirection angles.}
\begin{tabular}{|lll|c|}
\hline
Metric & LPIPS $\downarrow$ & Gaze Error $\downarrow$ \\
\hline
DeepWarp & 0.0946 & 14.18   \\
PRGAN & 0.0409 & 5.37  \\
CGR & 0.0565 & 9.19 \\
CFGR & {\bfseries 0.0333} & {\bfseries 5.15} \\
\hline
\end{tabular}
\label{tab:exp1}
\end{center}
\vspace{-0.5cm}
\end{table}

\begin{figure*}[htp]
\begin{center}
\includegraphics[width=1.0\linewidth]{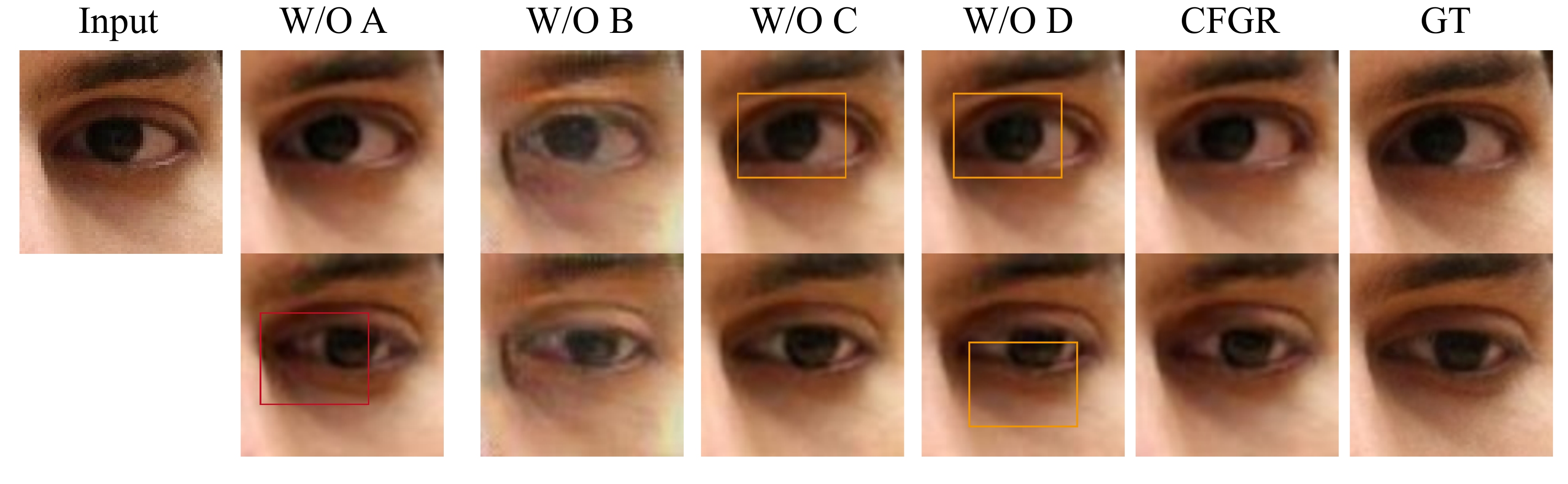}
\end{center}
\vspace{-0.5cm}
\caption{A qualitative comparison used in the ablation study. Red boxes: artifacts. Yellow boxes: unnatural shapes.}
\label{fig:exp2}
\vspace{-0.3cm}
\end{figure*}

\subsection{Results}
We first introduce the details of the qualitative and the quantitative evaluation protocols. For each head pose, we divide all the redirection angles into ten target groups by means of the sum of the direction differences in both pitch and yaw: $0^\circ$, $10^\circ$, $15^\circ$, $20^\circ$, $25^\circ$, $30^\circ$, $35^\circ$, $40^\circ$, $45^\circ$, $50^\circ$~(e.g., $0^\circ$ indicates that the angle differences between the target gaze and the input gaze are 0 in both the vertical and the horizontal direction). The test results of every group is used for the quantitative evaluation. Moreover, we select 10 redirection angles as the target angles for the qualitative evaluation: [$0^\circ$, $-15^\circ$], [$10^\circ$, $-15^\circ$], [$10^\circ$, $-10^\circ$], [$10^\circ$, $-5^\circ$], [$10^\circ$, $0^\circ$], [$10^\circ$, $5^\circ$], [$10^\circ$, $10^\circ$], [$10^\circ$, $15^\circ$], [$0^\circ$, $15^\circ$], [$-10^\circ$, $15^\circ$].

{\bfseries Qualitative Evaluation.}
In the 5th row of Fig.~\ref{fig:exp1_quan}, we show the redirection results of CFGR. The visually plausible results with respect to both the texture and the shape, and the high redirection precision, validate the effectiveness of the proposed model. Moreover,  compared to CGR (without the refined generator module), we note  that our refined model provides more detailed texture information and it eliminates unwanted artifacts and unnatural distortions of the iris shape.

\begin{figure*}[htp]
\begin{center}
\includegraphics[width=1.0\linewidth]{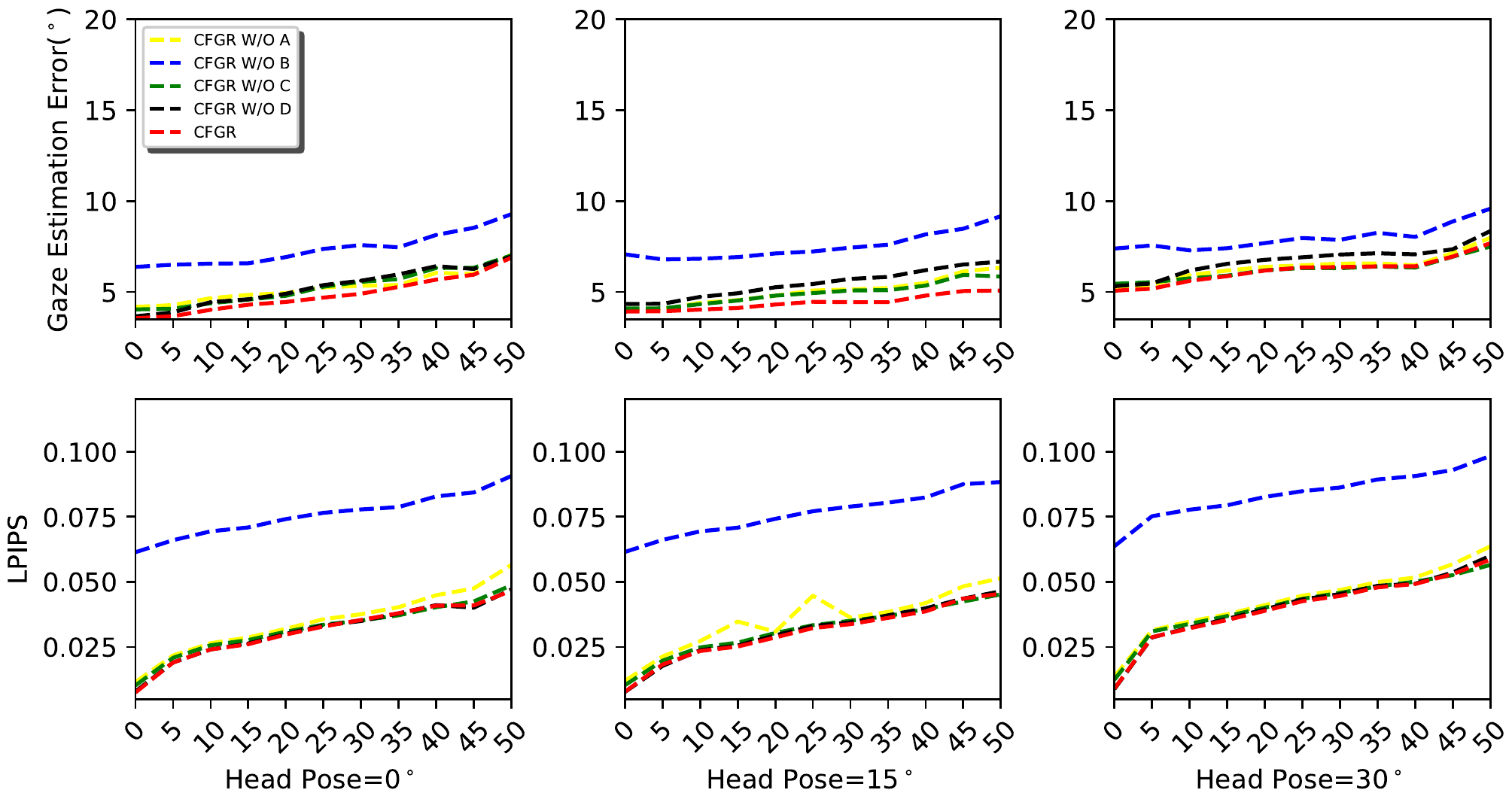}
\end{center}
\vspace{-0.5cm}
\caption{A quantitative evaluation used in the ablation study. The 1st row shows the gaze estimation error and the 2nd row the LPIPS scores. Our model ablated of the perceptual loss is called $A$, without the residual image learning is called $B$, without the flow field learning $C$, and without the pictorial gazemap guidance $D$.}
\label{fig:exp2_quan}
\vspace{-0.5cm}
\end{figure*}

As shown in the 2nd and in the 4th rows of Fig.~\ref{fig:exp1_quan}, we observe that both DeepWarp and CGR redirect the input gaze with respect to the target angles, which demonstrates the ability of flow field in representing the correct spatial transformation. However, DeepWarp has several obvious disadvantages~(marked with the yellow box in Fig.~\ref{fig:exp1_quan} and the corresponding zoom-in shown in the last row). For example, the generated textures are more blurry. In contrast, our coarse-grained CGR performs better. We attribute this to the fact that our encoder-decoder architecture with a bottleneck layer is better suitable for this task with respect to the scale-preserving  fully-convolutional architecture adopted in DeepWarp.

As shown in the 3rd and in the 5th row of Fig.~\ref{fig:exp1_quan}, both PRGAN and CFGR achieve high-quality redirection results with visual plausible textures and natural shape transformations for the iris. However, compared with CFGR, PRGAN suffers from two critical problems: (1) Lower image quality with a poor identity preservation~(marked with a red box on the left); (2) Incorrect redirection angles and blurry boundaries causing distortion of the eyeball~(marked with the yellow box and shown in the last row).

{\bfseries Quantitative Evaluation.}
In Fig.~\ref{fig:exp1},  we plot the gaze estimation errors and the LPIPS scores of different models. The three columns show the  redirection results with respect to  $0^\circ$,  $15^\circ$ and  $30^\circ$ head pose angle, respectively. Note that we combine the results of $\pm 15^\circ$ and $\pm 30^\circ$ head poses into $15^\circ$ and $30^\circ$. It can be observed from the 1st row of Fig.~\ref{fig:exp1} that CFGR achieves much lower gaze estimation error than DeepWarp and it is superior to PRGAN in most cases. Moreover, without the refined process, CGR has a much higher gaze error, especially for large gaze differences~(e.g., 50$^\circ$).

The 2nd row of Fig.~\ref{fig:exp1} shows the LPIPS scores. Here, we see that CFGR leads to much smaller scores than DeepWarp. Additionally, our model also has lower LPIPS scores than PRGAN, indicating that our method can generate a new eye image which is more perceptually similarly to the ground truth. However, CFGR has a higher gaze error or larger LPIPS scores in some cases, especially for redirection results with $30^\circ$ head pose. Overall, as shown in Table~\ref{tab:exp1}, our approach achieves 0.0333 LPIPS score, lower than the 0.0946 of DeepWarp, the 0.0409 of PRGAN and it gets a 5.15 gaze error, lower than the 14.18 of DeepWarp and the 5.37 of PRGAN.

{\bfseries User Study.}
We conducted a user study to evaluate the proposed model with respect to the human perception. In details, we divided the gaze redirection results on the test data into three groups with respect to the head pose of the input image and we randomly selected 20 samples generated by each method for each group. Then, for each image, 10 users were asked to indicate the gaze image that looks more similar with the ground truth. Table~\ref{tab:user_study} shows the results of this user study. We observe that our method outperforms PRGAN and DeepWarp in groups with $0^\circ$, $15^\circ$ and $30^\circ$ head poses. Moreover, CFGR is selected as the best model on average, as shown in the final column of Table~\ref{tab:user_study}.

\begin{table}
\begin{center}
\caption{Results of the user study using three different head poses (with ten generated samples per  pose). Every column sums to 100\%. The rightmost column shows the overall performance.} \label{tab:user_study}
\begin{tabular}{|lllll|c|}
\hline
Head Pose & 0$^\circ$ $\uparrow$ & 15$^\circ$ $\uparrow$ & 30$^\circ$ $\uparrow$ & Average $\uparrow$ \\
\hline
DeepWarp & 7.32\% & 10.18\% &  5.69\% & 7.73 \% \\
PRGAN & 30.12\% & 42.56 \%  & 45.79\%  & 39.49 \% \\
CFGR & {\bfseries 62.56\%} &  {\bfseries 47.26\%} & {\bfseries 48.52\%}  &  {\bfseries 52.78\%} \\
\hline
\end{tabular}
\vspace{-1.2cm}
\end{center}
\end{table}

\subsection{Ablation Study}
In this section we present an
 ablation study of the main components of our method. We  refer to the full model without the perceptual loss  as $A$. When we remove the flow learning in the encoder-decoder, this is called $B$. Removing the residual learning in the generator leads to model $C$, while removing the  gazemap pictorial guidance gets $D$ (more details below).

{\bfseries Perceptual Loss.}
Fig.~\ref{fig:exp2} shows that CFGR without the perceptual loss can generate results very close  to the full model. However, some of these results have more artifacts~(marked with a red box in the 2th column). Moreover, as shown in Fig.~\ref{fig:exp2_quan}, the gaze estimation error and the LPIPS score are larger when removing this  loss. Overall, the perceptual loss is helpful to slightly improve the visual quality and the redirection precision of the generated samples.

{\bfseries Residual Learning.}
We eliminate the residual term $R$ in Eq.~\ref{residual_learning} to evaluate its contribution. As shown in Fig.~\ref{fig:exp2}, the results are very blurry with a lot of artifacts. The quantitative evaluations in Fig.~\ref{fig:exp2_quan} are consistent with the qualitative results.

{\bfseries Flow Learning.}
Our encoder-decoder network predicts the flow field which is used to warp the input image for quickly learning the spatial shape transformation. As shown in Fig.~\ref{fig:exp2}, our full model achieves more natural results for the iris shape. Moreover, the quantitative results in Fig.~\ref{fig:exp2_quan} demonstrate the effectiveness of flow learning in improving the redirection precision.

{\bfseries Gazemap in NPG.}
When removing the gazemap~(see the 6th column in Fig.~\ref{fig:exp2}), the visual results present more shape distortions  compared with the full model. Moreover, the quantitative results in Fig.~\ref{fig:exp2_quan} demonstrate the effect of the gazemap in improving the redirection precision.

\section{Conclusion}
In this paper we  presented a novel gaze redirection approach based on a coarse-to-fine learning. Specifically, the encoder-decoder learns to warp the input image using the flow field for a coarse-grained gaze redirection. Then, the generator refines this coarse output by removing unwanted artifacts in the texture and possible distortions of the shape. Moreover, we proposed an NPG module which integrates a pictorial gazemap representation with the numerical angles to further improve the redirection precision. The qualitative and the quantitative evaluations validate the effectiveness of the proposed method and show that it outperforms the baselines with respect to both the visual quality and the redirection precision. In future work we plan to extend this approach to the gaze redirection task in the wild.

{\small
\bibliographystyle{ieee_fullname}
\bibliography{egbib}
}

\clearpage

\begin{figure*}
\vspace{-0.5cm}
\section{Appendix}
\subsection{More Gaze Redirection Results.}
\begin{center}
\includegraphics[width=0.9\linewidth]{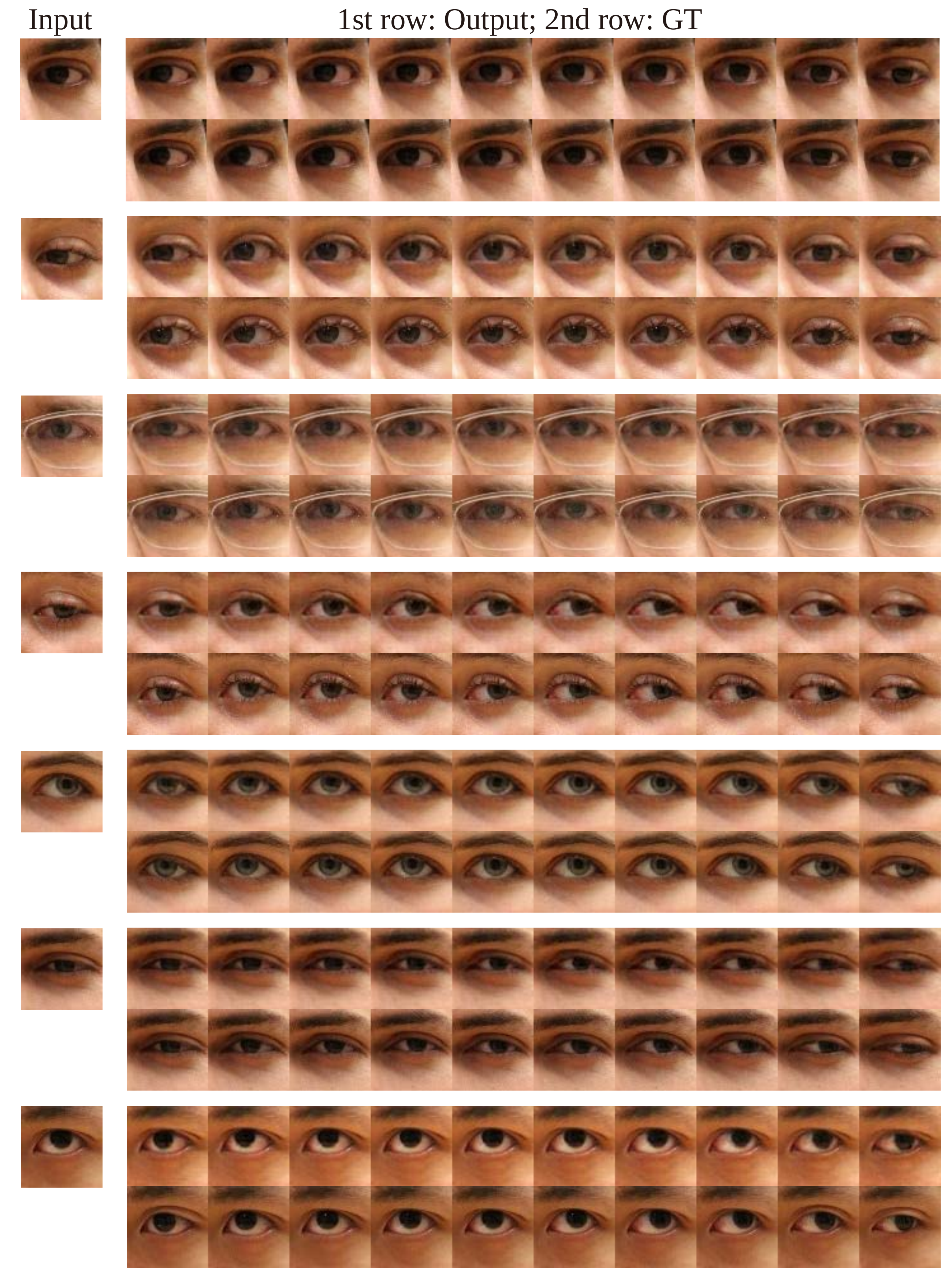}
\end{center}
\vspace{-0.3cm}
\caption{More high-quality gaze redirection results of CFGR}
\label{fig:supp2}
\end{figure*}

\begin{figure*}
\vspace{-0.5cm}
\section{More Gaze Redirection Results.}
\begin{center}
\includegraphics[width=0.9\linewidth]{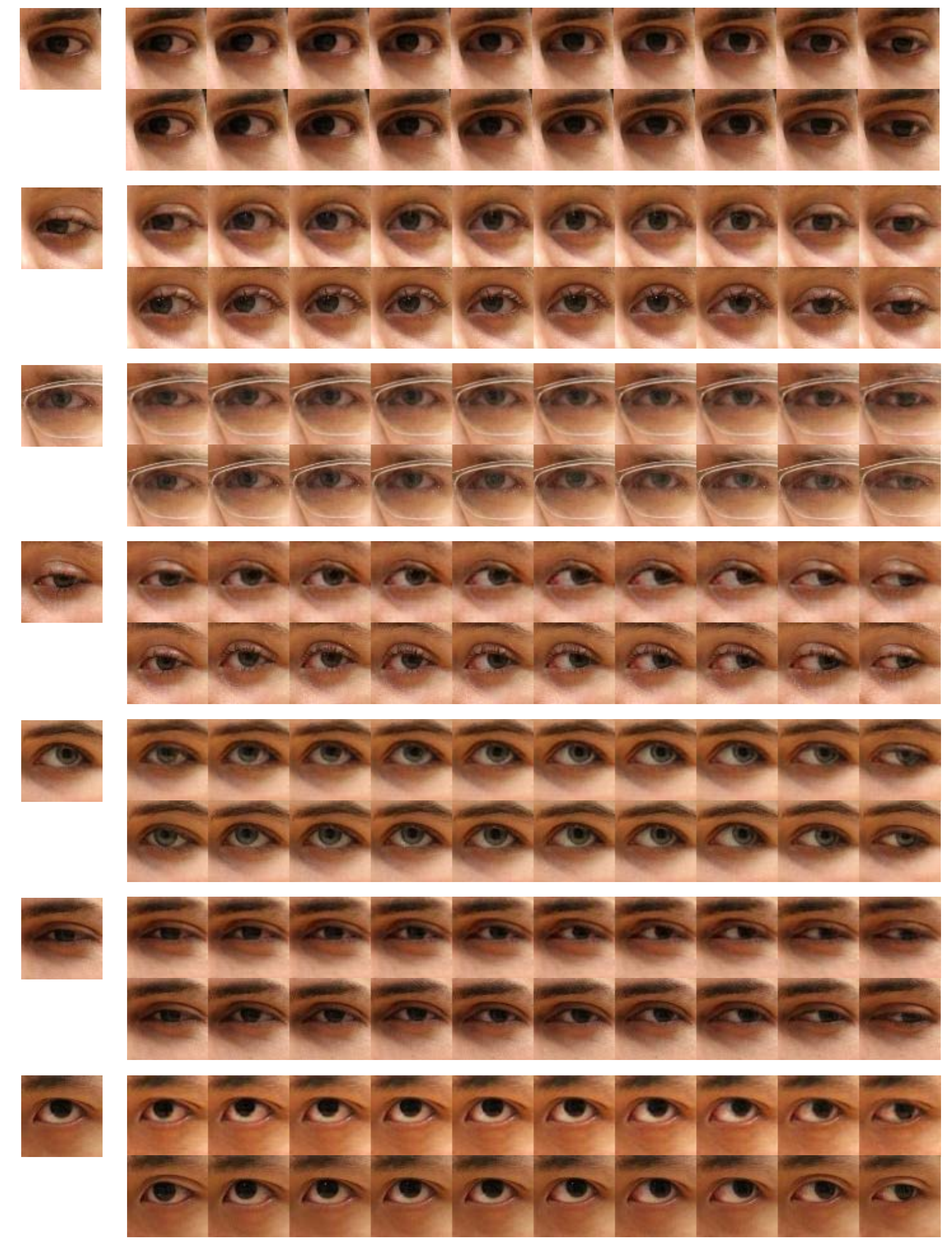}
\end{center}
\vspace{-0.3cm}
\caption{More high-quality gaze redirection results of CFGR}
\label{fig:supp3}
\end{figure*}

\begin{figure*}
\vspace{-0.5cm}
\section{More Binocular Gaze Redirection Results}
\begin{center}
\includegraphics[width=1.0\linewidth]{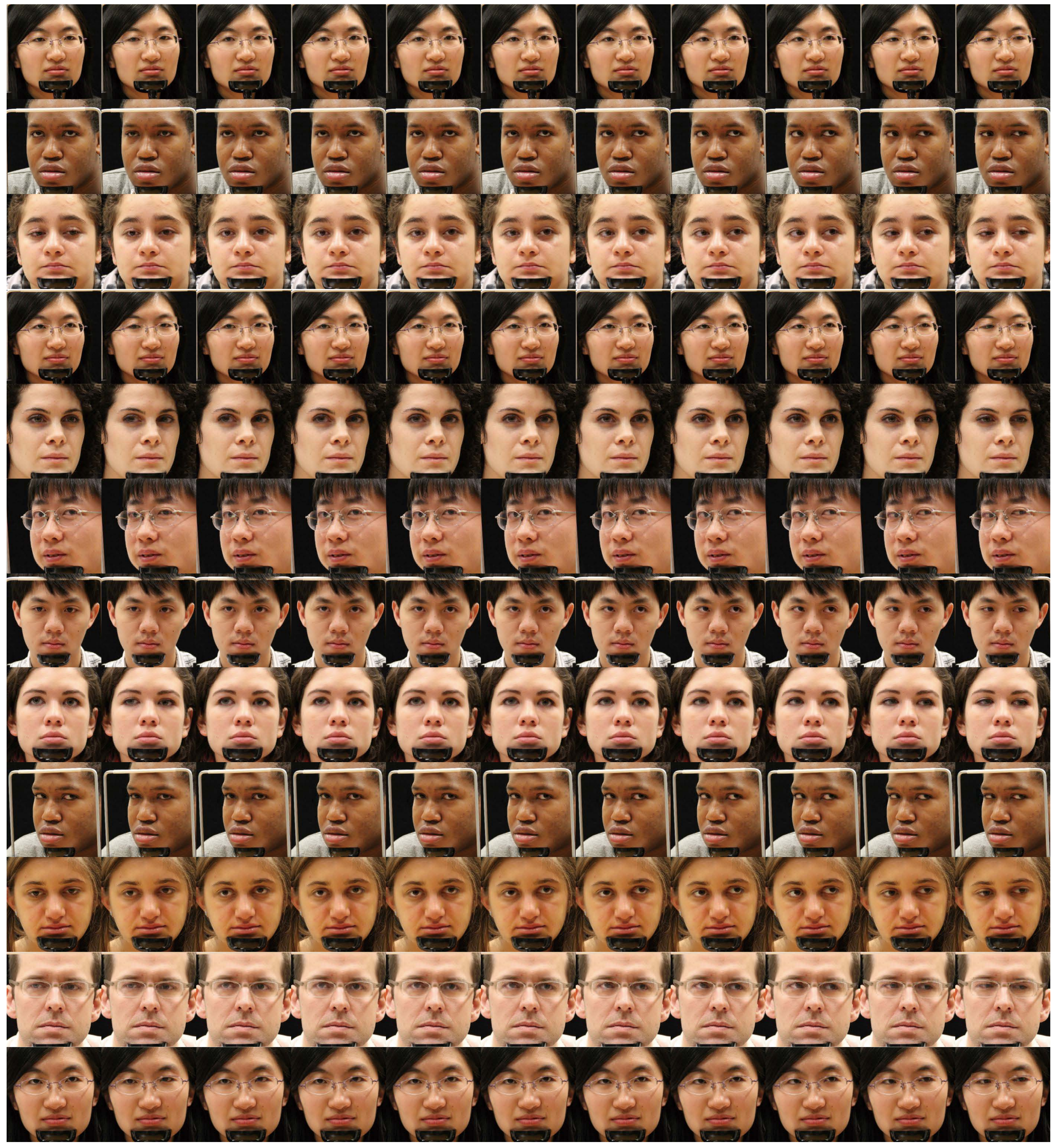}
\end{center}
\vspace{-0.3cm}
\caption{More high-quality binocular gaze redirection results of CFGR. For each row, the 1st column shows the input image, while the other columns show the results of the gaze redirection process with respect to different angles.}
\label{fig:supp4}
\end{figure*}

\end{document}